\documentclass[a4paper,fleqn]{cas-dc}

\usepackage[numbers]{natbib}
\usepackage{amsmath,amsfonts,amssymb}
\usepackage{multirow}

\newcommand\Tstrut{\rule{0pt}{3ex}}         
\usepackage[flushleft]{threeparttable}
\usepackage{balance}

\def\tsc#1{\csdef{#1}{\textsc{\lowercase{#1}}\xspace}}
\tsc{WGM}
\tsc{QE}
\tsc{EP}
\tsc{PMS}
\tsc{BEC}
\tsc{DE}

\begin{document}
\let\WriteBookmarks\relax
\def\floatpagepagefraction{1}
\def\textpagefraction{.001}
\shorttitle{Lightweight Framework for Tomato Leaf Disease Recognition}
\shortauthors{S Thuseethan et~al.}

\title [mode = title]{Siamese Network-based Lightweight Framework for Tomato Leaf Disease Recognition}

\author[cis]{Selvarajah Thuseethan}[orcid=0000-0001-6378-9940]
\ead{thuseethan@appsc.sab.ac.lk}

\author[cis]{Palanisamy Vigneshwaran}
\author[physical]{Joseph Charles}
\author[cis]{Chathrie Wimalasooriya}

\address[cis]{Department of Computing and Information Systems, Sabaragamuwa University of Sri Lanka, Belihuloya, Sri Lanka}
\address[physical]{Department of Physical Sciences and Technology, Sabaragamuwa University of Sri Lanka, Belihuloya, Sri Lanka}

\begin{abstract}
Automatic tomato disease recognition from leaf images is vital to avoid crop losses by applying control measures on time. Even though recent deep learning-based tomato disease recognition methods with classical training procedures showed promising recognition results, they demand large labelled data and involve expensive training. The traditional deep learning models proposed for tomato disease recognition also consume high memory and storage because of a high number of parameters. While lightweight networks overcome some of these issues to a certain extent, they continue to show low performance and struggle to handle imbalanced data. In this paper, a novel Siamese network-based lightweight framework is proposed for automatic tomato leaf disease recognition. This framework achieves the highest accuracy of 96.97\% on the tomato subset obtained from the PlantVillage dataset and 95.48\% on the Taiwan tomato leaf disease dataset. Experimental results further confirm that the proposed framework is effective with imbalanced and small data. The backbone deep network integrated with this framework is lightweight with approximately 2.9629 million trainable parameters, which is way lower than existing lightweight deep networks.
\end{abstract}
%
%


\maketitle

\section{Introduction}
\label{sec:introduction}
In the next 30 years, food production needs to be increased approximately by 60\% to feed the expected population of 10 billion people on the earth \cite{fao2019state}. Various natural and artificial activities significantly reduce food production while anticipating to reach the expected level. The hike in food production with potential losses should meet the global demand for food. In general, plant diseases are a major threat to the modern agricultural industry, heavily reducing the production and food quality \cite{strange2005plant}. The crop pests and diseases cause a considerable loss of global yields, approximately 21.5\% and 17.2\% in wheat and potato plants, respectively \cite{savary2019global}.

Tomato is one of the most economically and nutritionally essential vegetables, which is cultivated all around the globe. The worldwide tomato production has already exceeded 187 million tonnes by 2020, with China being the largest producer, followed by India and the United States of America \cite{panno2021review}. Tomato crops can be easily affected by numerous diseases that can cause dramatic economic losses and food shortages. Hence, farmers need to diagnose tomato diseases as early as possible to reduce the risk of losing yields. Apart from preventing the tomato diseases by annually testing the garden soil and maintaining an adequate level of potassium, early days, farmers used manual inspection of affected leaves to identify the tomato diseases \cite{eligar2022performance}. However, the manual diagnosis of tomato disease is a tedious and time-consuming practice, requiring plant disease-related expertise.  

Recently proposed computational methods to automatically recognize tomato diseases from leaf images used both handcrafted \cite{vadivel2022automatic,zhang2021rapid} and deep learning \cite{eligar2022performance,kim2021tomato,liu2020early,yoren2021tomato,zhang2020deep} based feature extraction techniques. The handcrafted feature extraction is a process of learning the most informative features to train the classifiers for recognition tasks. The methods adapting the handcrafted feature extraction scheme contain multiple steps in the pipeline, such as pre-processing, feature extraction and classification. Even though the handcrafted feature extraction techniques are faster in comparison to deep learning methods, they often showed poor performance in unfamiliar conditions. Additionally, the handcrafted techniques are generally not end-to-end trainable, increasing the intricacy of the overall tomato disease recognition system.

On the other hand, very recently proposed deep learning-based image classification techniques showed promising performance across several application domains \cite{lecun2015deep}. Medical, affective computing, social and agriculture are some popular domains where image classification techniques are extensively utilized. Similar to the other image classification tasks, the automated feature extraction-based deep learning techniques have become the benchmarks in tomato disease recognition \cite{eligar2022performance}. Researchers not only utilized the state-of-the-art deep networks \cite{chen2022alexnet} but also proposed tailor-made deep networks \cite{yoren2021tomato,kim2021tomato} for tomato disease recognition. The majority of the traditional training process of the deep networks often demands high computational requirements and large labelled data for training as they use very deep networks. Further, they are not feasible for real-time solutions as they consume a considerable amount of time in both training and inference phases.

As a potential solution, the lightweight deep networks have been widely used for tomato disease recognition, which facilitates smooth mobile deployment and training with small data \cite{wu2020lightweight}. However, lightweight network-based tomato disease recognition methods often show limited performance with real-world samples, which is one of the major limitations that hinder its widespread applicability. To enhance the performance, they predominantly rely on additional mechanisms like an attention scheme that further complicate the resultant model \cite{bhujel2022lightweight}. Commonly, the lightweight networks with traditional training schemes suffer in obtaining expected classification accuracy with imbalanced data. 

To address these issues, in this paper, a novel Siamese network-based lightweight tomato disease recognition framework is proposed. Figure \ref{fig:overallarchitecture} illustrates the overall architecture of the proposed tomato disease recognition framework. By adopting the key characteristics of the Siamese networks, this lightweight framework can effectively recognize tomato diseases using imbalanced and small data. The key contributions of this paper are given as follows:

\begin{figure*}
	\centering
	\includegraphics[width=1\linewidth,scale=1.5]{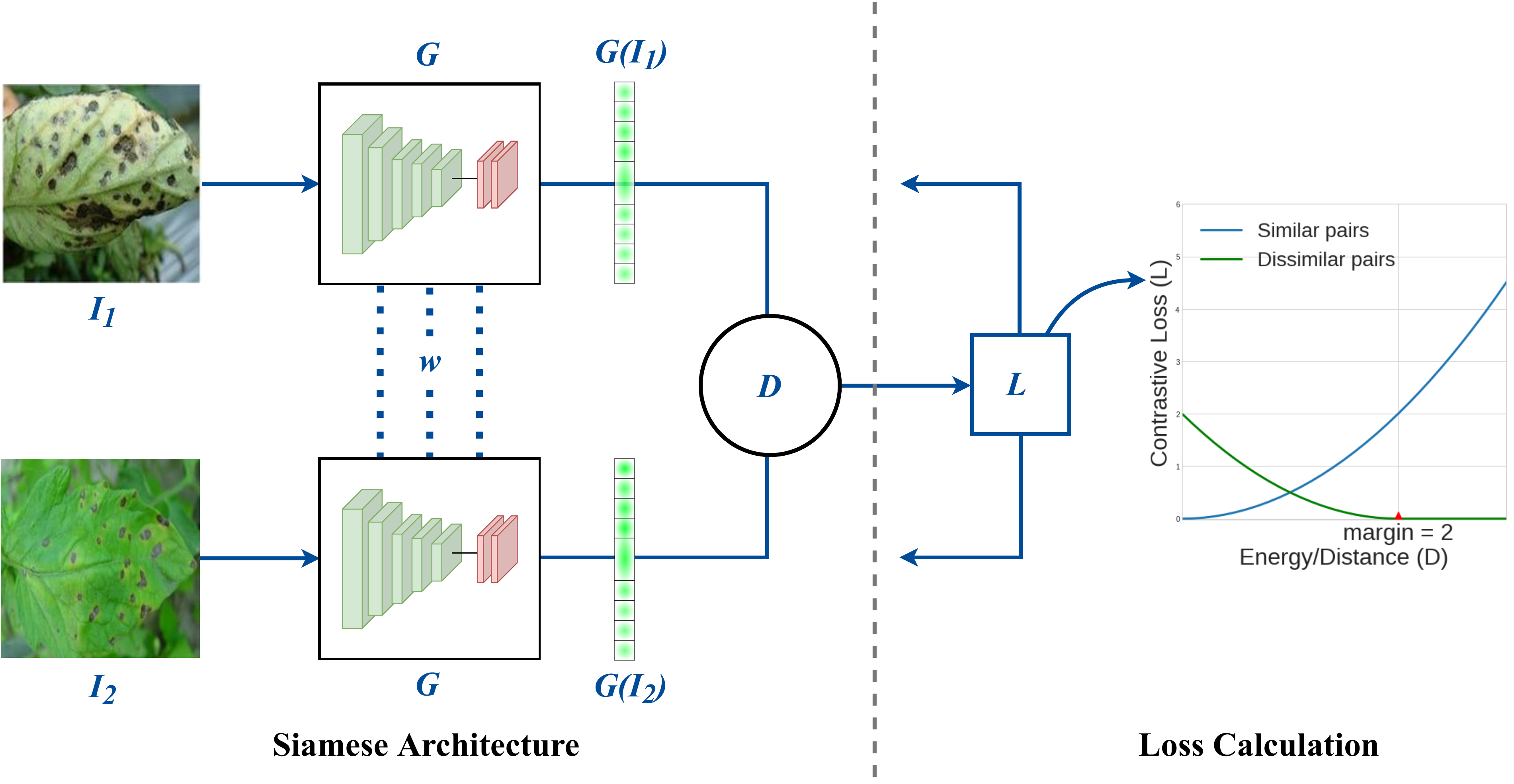}
	\caption{The overall architecture of the proposed tomato disease recognition framework. The $I_1$ and $I_2$ are the input images. The weights $w$ are shared between two streams ($G$) of the Siamese network. Figure \ref{fig:backbone_network} illustrates the architecture of the backbone network $G$. The distance $D$ is estimated as the euclidean distance between the outputs $G(I_1)$ and $G(I_2)$. The contrastive loss $L$ is then calculated based on $D$.}
	\label{fig:overallarchitecture}
\end{figure*}

\begin{enumerate}
	\item A novel Siamese network-based lightweight framework with a contrastive loss is proposed to recognize tomato diseases from small and imbalanced leaf data.
	
	\item A testing procedure dedicated for Siamese networks using the majority voting is proposed to enhance the tomato disease recognition accuracy.
	
	\item A set of extensive experiments were carried out on two tomato leaf datasets\footnote{https://data.mendeley.com/datasets/ngdgg79rzb/1}, including the tomato subset of the PlantVillage dataset and the Taiwan tomato leaves dataset, to show the feasibility of the proposed framework in effectively recognising diseases.
	
	\item The comparative study shows that the proposed framework can show better performances with small and imbalanced data, in comparison to existing tomato disease recognition from leaf images.
	
\end{enumerate}

The rest of the paper is organized as follows. A comprehensive analysis of the tomato leaf disease recognition is provided in Section \ref{sec:relatedwork}. Next, the proposed Siamese network-based framework is presented in Section \ref{sec:proposedmethod}. A set of experimental results are discussed in Section \ref{sec:experiments}. Section \ref{sec:conclusion} concludes the paper with a few future directions on automated tomato leaf disease recognition. 

\section{Related Work}
\label{sec:relatedwork}
Diagnosing plant diseases on time in an automatic manner by utilising the latest information processing techniques has been actively researched lately in pursuit of supporting the agricultural industry to address its emerging challenges \cite{ngugi2021recent}. Over the years, the domain experts in agriculture considered effective recognition of diseases for different crop types, such as citrus \cite{janarthan2020deep} and apple \cite{yu2022apple}. Despite the fact that contemporary research has primarily focused on advanced deep learning techniques, in this section, some of the most significant tomato disease recognition techniques covering both traditional and deep learning approaches are reviewed. However, readers are advised to read the recent surveys, \cite{hasan2020review} and \cite{li2021plant}, to acquire comprehensive knowledge about more generic plant disease recognition techniques.

\begin{figure*}
	\centering
	\includegraphics[width=1\linewidth,scale=1.5]{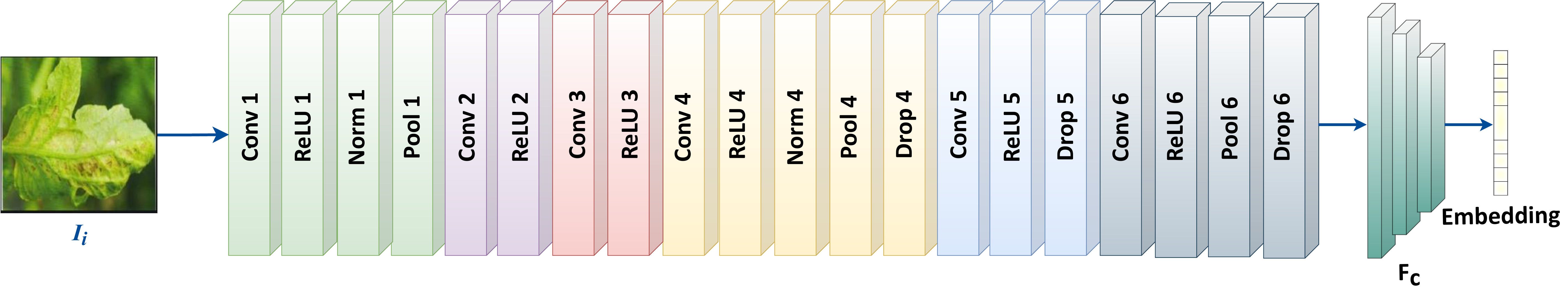}
	\caption{The block diagram of the lightweight deep network used as the backbone of the proposed framework. The layer configuration details are further given in Table \ref{tab:network}}
	\label{fig:backbone_network}
\end{figure*}

Traditionally, similar to image classification tasks in other domains, handcrafted feature extraction schemes are most widely adopted in tomato disease classification using leaf images \cite{vadivel2022automatic,zhang2021rapid}. For example, in \cite{zhang2021rapid}, the colour and texture features, along with the kernel mutual subspace method (KMSM), are used for rapid recognition of tomato disease stages. This method achieved an accuracy of 99.34\% and 98.66\% on a subset of the PlantVillage dataset and AI Challenger 2018 dataset. Further, the proposed method is faster than VGG16 and support vector machine (SVM) in terms of average training and recognition time. Further, various kinds of handcrafted feature extraction techniques have been used for plant disease recognition in general \cite{singh2022hybrid}. However, as the handcrafted feature extraction scheme is a manual process of extracting important features, the methods that use this scheme must follow a series of phases, such as image pre-processing (including segmentation/cropping), feature extraction and classification. This pipeline not only increases the complexness of the tomato disease recognition method but also restricts the end-to-end train-ability.

Recent advancements in deep learning for present-day pattern recognition tasks across different domains resulted in creating better opportunities in plant disease recognition \cite{li2021plant}. Several deep learning-based recently proposed showed promising results in tomato disease recognition. For example, in an early work, pre-trained AlexNet and VGG16 deep networks are utilized to classify crop diseases from tomato leaves and validated on tomato leaf images taken from the PlantVillage dataset \cite{rangarajan2018tomato}. Both AlexNet and VGG16 deep networks consist of a large number of trainable parameters, and thus they are deemed to be very heavy and demand large training data. Chen et. al. \cite{chen2020identification} proposed a novel Both-channel Residual Attention Network (B-ARNet) model for automated tomato leaf disease recognition. In this study, image denoising and enhancing processes along with background removal are performed prior to extracting the features using the B-ARNet model, which is not ideal for end-to-end system implementation. Unlike the methods discussed so far, a simple tailor-made deep convolutional neural network (CNN) is proposed in \cite{agarwal2020toled} for accurate tomato leaf disease detection. However, it shows comparatively poor performance with the lowest accuracy of around 76\%. Furthermore, the use of attention schemes has proven to improve the classification accuracy of CNN-based tomato disease recognition models \cite{zhao2021tomato}. Attaching additional modules to deep networks often caused increased trainable parameters.

The lightweight deep networks are well known for their reduced memory consumption and real-time applicability. Some lightweight networks like MobileNets provide the facility to find the trade-off between accuracy and latency with the help of tunable hyperparameters \cite{howard2017mobilenets}. Most of the recent plant disease recognition methods have been aimed at developing lightweight networks because farmers predominantly expect in-the-wild applicability \cite{thakur2022trends}. In \cite{zaki2020classification}, a pre-trained MobileNetv2 network is fine-tuned for a three-class tomato disease recognition task. Although a considerable overall accuracy (>90\%) is achieved, the proposed method is validated with a simple three-class tomato disease dataset. Bhujel et. al. \cite{bhujel2022lightweight} proposed a lightweight model that shows an increased overall accuracy and reduced network parameters. In order to enhance the performance of the proposed lightweight model, an attention module is embedded in the original network. One of the key challenges of lightweight architectures is they frequently show limited performance due to their low generalization capability. Further, the classical training of a network regardless of whether it is heavy or lightweight not only demands a considerable amount of training data but also fails to cope with imbalanced data.

In summary, the majority of the existing tomato leaf disease recognition methods are not feasible to be adopted in real-time settings due to practical problems, such as the unavailability of large and balanced data and the limited computational power and memory. The lightweight Siamese architectures in recent years have shown immaculate performances with small and imbalanced data \cite{bromley1993signature}. Hence, there is a necessity to develop a fully-fledged framework by realizing the advantages of Siamese architecture as the first step toward addressing these challenges.

\section{Proposed Method}
\label{sec:proposedmethod}
In this section, the proposed lightweight tomato disease recognition framework using the Siamese network is comprehensively elaborated. First, the overall architecture of the novel Siamese network given in Figure \ref{fig:overallarchitecture} is explained. Second, the loss calculation procedure is elaborated. Third, the newly constructed lightweight CNN that serves as the backbone is given in detail. Finally, the proposed novel testing procedure is presented.

\subsection{Overall Siamese Architecture}
Figure \ref{fig:overallarchitecture} provides the overview of the proposed Siamese network-based lightweight tomato leaf disease recognition framework along with the contrastive loss calculation procedure. The in-depth configuration details of the sub-networks (backbone networks) are suppressed for brevity. In general, a Siamese architecture is a special kind of neural network comprising two or more streams of identical backbone networks \cite{bromley1993signature}. In Siamese architectures, both identical sub-networks are constructed with the same parameter sets and network configuration. The proposed Siamese framework consists of two streams of sub-networks that take two different tomato leaf images as inputs at a time. 

The output patterns (i.e., feature embeddings) obtained from both sub-networks are compared to find out the similarity measure (i.e., the distance), which is then used in the decision-making process. The distance value is based on the pairwise euclidean distance between the output feature vectors obtained by both streams of the Siamese network. In mathematics, the length of a line segment connecting two points in euclidean space is referred to as the euclidean distance, which is the most common distance measure used in machine learning. Ideally, the distance is kept $0$ for similar pair of images and a value $>0$ for dissimilar pair of images. For example, the distance is $0$ for both of the images taken from Bacterial spot disease and a value $>0$ for one image taken from Bacterial spot disease and the other taken from Black mold disease. During the training phase, for each pair of images, $I_1$ and $I_2$, the euclidean distance $D$ is derived through a parameterized distance function $D(I_1, I_2)$ defined as:
\begin{equation} \label{eq:euclideandistance}
	D(I_1, I_2) = \lVert G(I_1) - G(I_2) \rVert_2
\end{equation}
where, euclidean distance $D(I_1, I_2)$ is subsequently used as the energy in loss calculation. In the rest of the paper, the euclidean distance $D(I_1, I_2)$ is referred to as $D$ for conciseness.

\subsection{Loss Calculation}
In this method, a widely popular contrastive loss proposed in \cite{hadsell2006dimensionality} is exploited. The contrastive loss calculation is performed over the pairs of input samples, in contrast to the conventional method, where the loss is calculated as the sum over input samples. In the contrastive loss, the similar input pairs are pushed close by while the dissimilar input samples are pulled apart. The contrastive loss calculation for the $i^{th}$ pair of input images is given in Eq. \ref{eq:contrastive}.
\begin{equation} \label{eq:contrastive}
	L (Y, [I_1, I_2]^i) = (1-Y)L_s(D^i) + (Y) L_d(D^i)
\end{equation}
where, $L (Y, [I_1, I_2]^i)$ is the loss for the $i^{th}$ image pair $[I_1, I_2]^i$ with the label $Y$. The partial loss for similar pairs $L_s$ and dissimilar pairs $L_d$ are designed in such a way that the overall loss $L$ is minimized. Correspondingly, the $D$ fetch a small value for similar input pairs and a large value for dissimilar input pairs. The resultant contrastive loss is given below.
\begin{equation} \label{eq:regionlandmarks}
	L = \underbrace{(1-Y)\frac{1}{2}(D)^2}_{\substack{\text{similar pair loss}}} + \underbrace{(Y) \frac{1}{2}\{max(0, m - D)\}^2}_{\substack{\text{dissimilar pair loss}}}
\end{equation}
here, $Y$ indicates a binary label given to an input image pair $[I_1, I_2]$. For similar and dissimilar input pairs, $Y$ has the values $0$ and $1$, respectively. $m$ is the margin set to a value greater than $0$. For dissimilar pairs, any distance $D$ within the margin $m$ (i.e., $m-D>0$) is penalized. In this case, the margin $m$ is set to $2$ defines the loss tendency of dissimilar pairs, as shown in the loss calculation fragment of Figure \ref{fig:overallarchitecture}. Likewise, similar pairs contribute to the contrastive loss if the distance $D>0$. 

\begin{table}[t]
	\begin{center}
		\caption{
			Configurations of the backbone CNN architecture integrated with the proposed Siamese network. In this table, $w$ = width, $h$ = height, $c$=channels, $f$ = number of filters (output channels), $ks$ = kernel size, $s$ = stride, $p$ = padding, $sz$ = amount of neighbouring channels, $\alpha$ = multiplicative factor, $\beta$ = exponent, $af$ = additive factor, $pr$ = probability of an element to be zeroed, $i$ = size of each input and $o$ = size of each output
		}
		\begin{tabular}{|p{0.6cm}|l|l|}
			\hline
			\textbf{Block} &\textbf{Layers}&\textbf{Configuration\Tstrut} \\
			\hline
			0 & Input				& $w$=128, $h$=128, $c$=3 \Tstrut\\
			1 & Conv 1				& $c$=3, $f$=64, $ks$=5, $s$=1, $p$=1 \\
			& ReLU 1 				& \\
			& Norm 1			 	& $sz$=5 , $\alpha$=0.0001 , $\beta$=0.75 , $af$=2 \\
			& Pool 1 				& $ks$=3 , $s$=2 \\
			2 & Conv 2				& $c$=64, $f$=96, $ks$=3, $s$=1, $p$=2 \\
			& ReLU 2 				& \\
			3 & Conv 3				& $c$=96, $f$=128, $ks$=3, $s$=1, $p$=2 \\
			& ReLU 3				& \\
			4 & Conv 4				& $c$=128, $f$=96, $ks$=3, $s$=1, $p$=2 \\
			& ReLU 4 				& \\
			& Norm 4			 	& $sz$=5 , $\alpha$=0.0001 , $\beta$=0.75 , $af$=2 \\
			& Pool 4 				& $ks$=3 , $s$=2 \\
			& Drop 4	 			& $pr$=0.2 \\
			5 & Conv 5				& $c$=96, $f$=64, $ks$=1, $s$=1, $p$=1 \\
			& ReLU 5 				& \\
			& Drop 5 				& $pr$=0.2 \\
			6 & Conv 6				& $c$=64, $f$=32, $ks$=1, $s$=1, $p$=1 \\
			& ReLU 6 				& \\
			& Pool 6 				& $ks$=3 , $s$=2 \\
			& Drop 6 				& $pr$=0.2 \\
			
			7 & FC 7				& $i$=32$\times$324, $o$=256\\
			& ReLU 7 				& \\
			& Drop 7 				& $pr$=0.5 \\
			8 & FC 8				& $i$=256 , $o$=64\\
			& ReLU 8 				& \\
			9 & FC 9				& $i$=64 , $o$=32\\
			\hline 
		\end{tabular}
		\label{tab:network}
	\end{center}
	
\end{table}

\subsection{Lightweight Backbone Network}
In recent years, deep networks with fewer layers have become very popular in the image classification domain due to their lightweight nature. As the primary focus of this study is to develop the siamese framework lighter, the number of layers is retained to a small value, and so forth the trainable parameters of this network are $\approx2.96$ millions. As described in Figure \ref{fig:backbone_network} and Table \ref{tab:network}, a nine-block deep CNN is proposed as the backbone network. The in-depth details about the network configurations of the backbone CNN architecture are provided in Table \ref{tab:network}.

The images in the size of $3\times128\times128$ (i.e., channel $\times$ width $\times$ height) can be fed as the inputs, as given in Table \ref{tab:network}. This network consists of six convolution layers and three connected layers to learn the most important feature representations. The receptive fields of the convolution layers are kept very small ($5\times5$, $3\times3$ and $1\times1$). More importantly, the receptive field of $1\times1$ used in convolution layers $5$ and $6$ portrays the linear transformation function that improves the non-linear property of the neural network. The rectified linear (ReLU) layers are stacked after each convolution layer (i.e., convolution layers $1$-$6$) and after the first two fully connected layers (i.e., fully connected layers $7$ and $8$), as they also consistently increase the non-linearity.

The local response normalisation (LRM) function, referred as Norm $1$ and $4$ in Table \ref{tab:network}, stimulate lateral inhibition that performs local contrast enhancement. Two LRM layers are appropriately affixed in blocks $1$ and $4$ immediately after ReLU layers. Further, 2-dimensional max-pooling layers (indicated as Pool in Table \ref{tab:network}) that estimate each patch's maximum value in every feature map are placed in blocks $1$, $4$ and $6$. In blocks $4$, $5$, $6$ and $7$, three dropout layers with probability retention of $0.2$, $0.2$, $0.2$ and $0.5$ are stacked, respectively. The final layer of the backbone CNN architecture is a fully connected layer with $32$ neurons.

\subsection{Majority Voting-based Testing Scheme}
\begin{figure*}
	\centering
	\includegraphics[width=1\linewidth,scale=1.5]{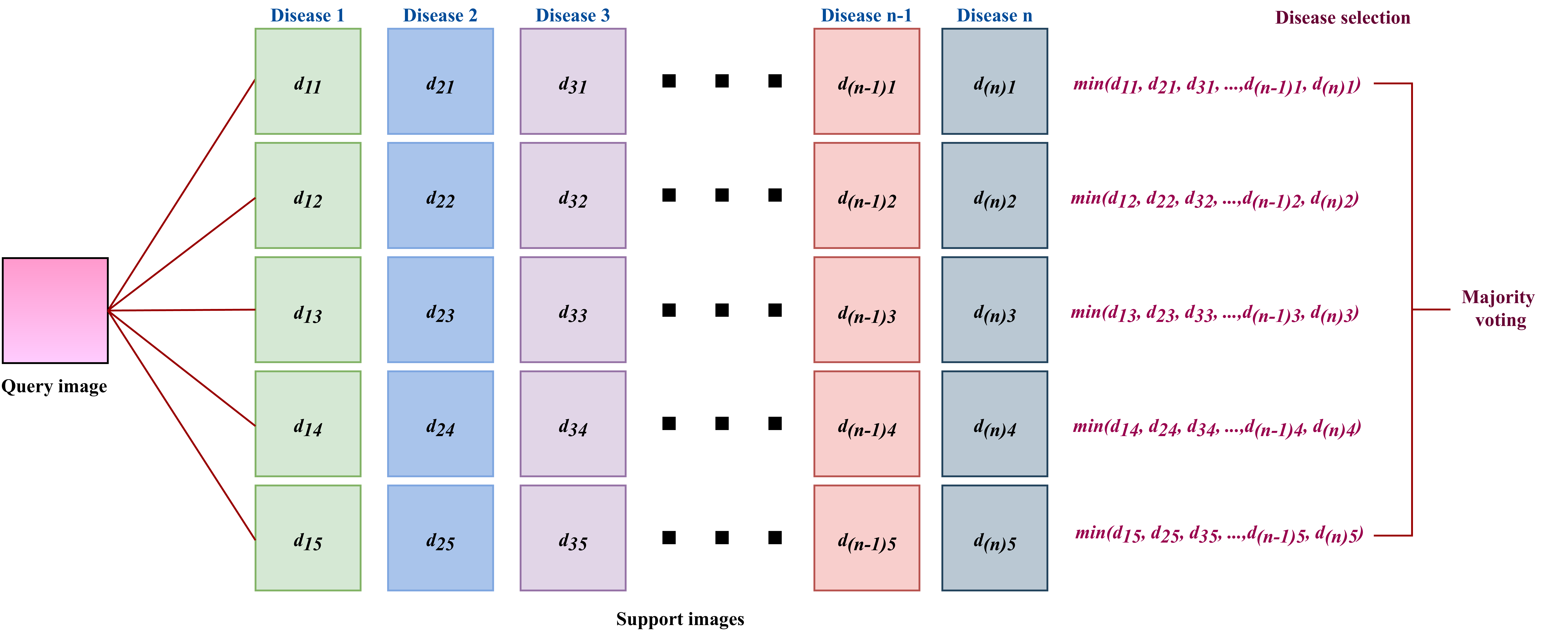}
	\caption{Proposed testing scheme dedicated for the Siamese network-based tomato plant disease recognition. This scheme follows a majority voting mechanism.}
	\label{fig:testing_framework}
\end{figure*}

In order to obtain the testing accuracy of the proposed tomato disease recognition approach, a querying mechanism involving support images and the query image is exploited, as illustrated in Figure \ref{fig:testing_framework}. 

Consider that there are $n$ tomato disease types ($Disease$ $1$ to $n$) to be classified. For each disease, five support images are arbitrarily selected. The distance $d_{i,j}$ is estimated between the query image and the $j^{th}$ support image of disease $i$, where $j$ takes $1$ to $5$ and $i$ takes $1$ to $n$. The disease type of the image with minimum distance amongst a row of support images is selected as the predicted disease class, which is given by:
\begin{equation} \label{eq:disease}
	Disease^j = f_{disease}(\min_{i = 1 \dots n} (d_{ij}))
\end{equation}
where, $j$ is a constant that will take the values between $1$ and $5$ for multiple runs of this equation. The $f_{disease}(\cdot)$ is a function that returns the disease type for the input given. For example, the $Disease^1 = Disease (n-1)$ if the $d_{(n-1)1}$ shows the minimum distance for $j=1$.

Once the $Disease^j$ for all $j = 1\dots5$ are decided, a majority voting scheme is applied to make the final decision. To perform this, first, the occurrences of each tomato leaf disease type selected within $Disease^j$ are counted. Next, the tomato disease type with the highest occurrence count is selected as the predicted tomato disease class. For instance, the tomato leaf $Disease 3$ can be fixed as the predicted class if it is selected $3$ out of $5$ times. In the case of two or more equal number of maximum occurrences, the average distances of the disease types with maximum occurrences are used as the deciding factor. Precisely, the predicted tomato disease type is selected as the one with a minimum average distance from the maximum occurrents.

\section{Experiments}
\label{sec:experiments}
In this section, the results obtained for a set of experiments performed in various environments are discussed in detail. The primary focus of these experiments is to demonstrate the feasibility of the proposed model with small and imbalanced data. First, the tomato disease datasets are described, followed by implementation details. Next, the results are discussed in detail, along with the comparison of existing state-of-the-art methods.

\subsection{Datasets}
\begin{figure*}
	\centering
	\includegraphics[width=1\linewidth,scale=1.5]{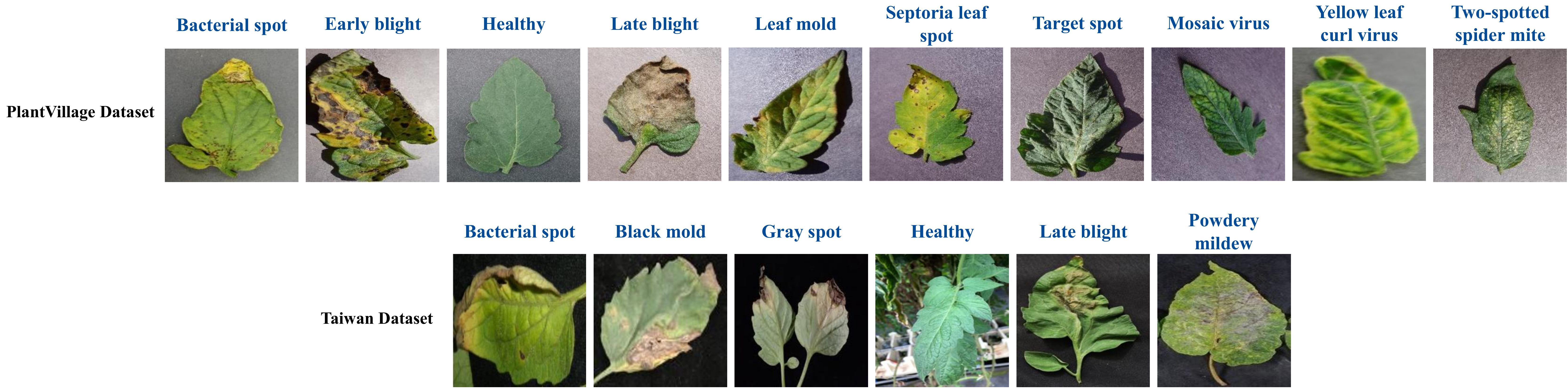}
	\caption{Sample images taken from the PlantVillage and Taiwan datasets for each tomato disease class.}
	\label{fig:datasets}
\end{figure*}

To evaluate the proposed lightweight tomato leaf disease recognition framework, two benchmark datasets, such as a subset of the PlantVillage dataset containing tomato leaf disease samples and the Taiwan tomato disease dataset, are used \cite{huang2020dataset}. Sample images picked from each class of both datasets are given in Figure \ref{fig:datasets}.

The \textit{PlantVillage tomato subset} consists of $14,531$ image samples for ten classes, including nine tomato leaf diseases classes and one healthy class. The tomato leaf diseases included in this dataset are Bacterial spot ($1702$), Early blight ($800$), Late blight ($1528$), Leaf mold ($762$), Septoria leaf spot ($1417$), Target Spot ($1124$), Mosaic virus ($299$), Yellow leaf curl virus ($4286$) and Two-spotted spider mite ($1341$). This is a highly imbalanced dataset with uneven distribution of image samples, where the majority class ($4286$ samples of Yellow leaf curl virus disease class) consists of fourteen times ($14\times$) larger sample size than the minority class ($299$ of Mosaic virus disease class). The image size of the original dataset is $227\times227$.

\begin{figure}
	\centering
	\includegraphics[width=1\linewidth,scale=1.5]{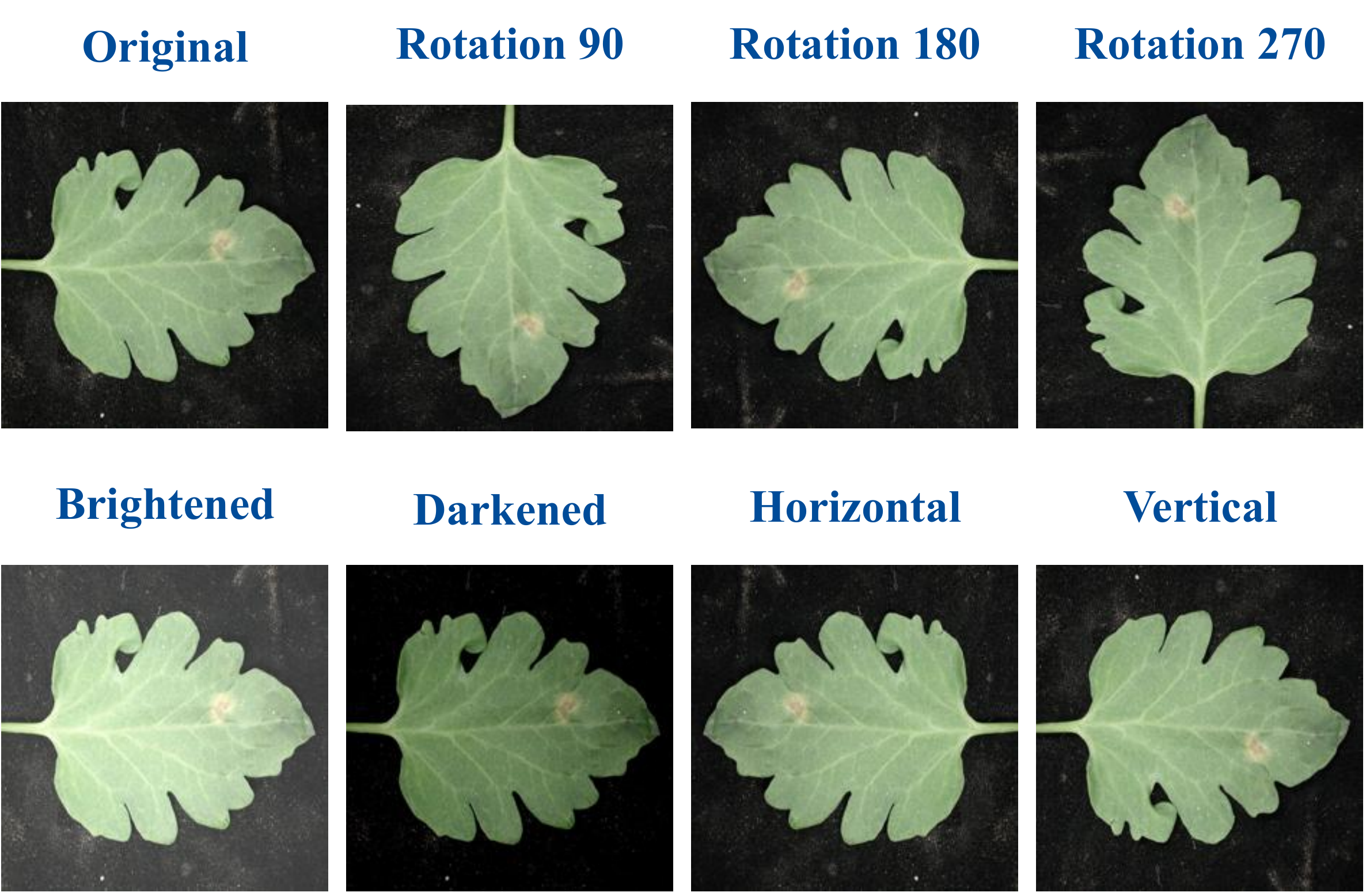}
	\caption{An example image taken from the Black mold disease and its augmented samples. There are $7$ different augmentation functions are applied on the leaf images.}
	\label{fig:augmentation}
\end{figure}

The original \textit{Taiwan tomato disease dataset} is relatively a small one used to evaluate the proposed model, which has $622$ leaf samples in the size of $227\times227$ for six categories, including healthy and diseased classes. The Bacterial spot ($100$), Black mold ($67$), Gray spot ($84$), Healthy ($106$), Late blight ($98$) and Powdery mildew ($157$) are the classes included in this dataset. Furthermore, there is an enhanced dataset containing more samples for each class that are generated using a range of augmentation functions, such as clockwise rotation (by 90, 180 and 270 degrees), mirroring (horizontally and vertically) and changing image brightness, which is also exploited. Figure \ref{fig:augmentation} provides an example image with all the augmented samples. There are $4,976$ tomato leaf images in the final dataset with augmented samples.

\subsection{Implementations and Evaluation Metrics}
The proposed framework and the other compared methods are implemented using the PyTorch\footnote{https://pytorch.org/} open source machine learning framework and trained on Google Colab\footnote{https://colab.research.google.com/} notebook. The models are evaluated on Intel(R) Xeon(R) CPU @ 2.30GHz supported by NVIDIA Tesla T4 GPU with 13GB RAM. 

All the models are trained for $10$ epochs with a batch size of $8$. In the training phase, the initial learning rate of the Adam optimizer is set to $0.001$ with a weight decay of $0.0001$. Further, the results on the Taiwan tomato disease dataset are recorded for the dedicated test set provided by the authors, while the results on the PlantVillage tomato subset are presented using a 10-fold cross-validation technique. To present the experimental results, a popular evaluation metric average accuracy, is used. The closeness of the predicted tomato disease class to the true tomato disease class can be effectively described through accuracy. The equation to obtain accuracy is defined below. 
\begin{equation} \label{eq:accuracy}
	Accuracy = \frac{t_p + t_n}{t_p + t_n + f_p + f_n}
\end{equation}
where, $t_p$ is true positive, $t_n$ is true negative, $f_p$ is false positive and $f_n$ is false negative.

\subsection{Results}
\begin{figure}
	\centering
	\includegraphics[width=0.9\linewidth,scale=1.5]{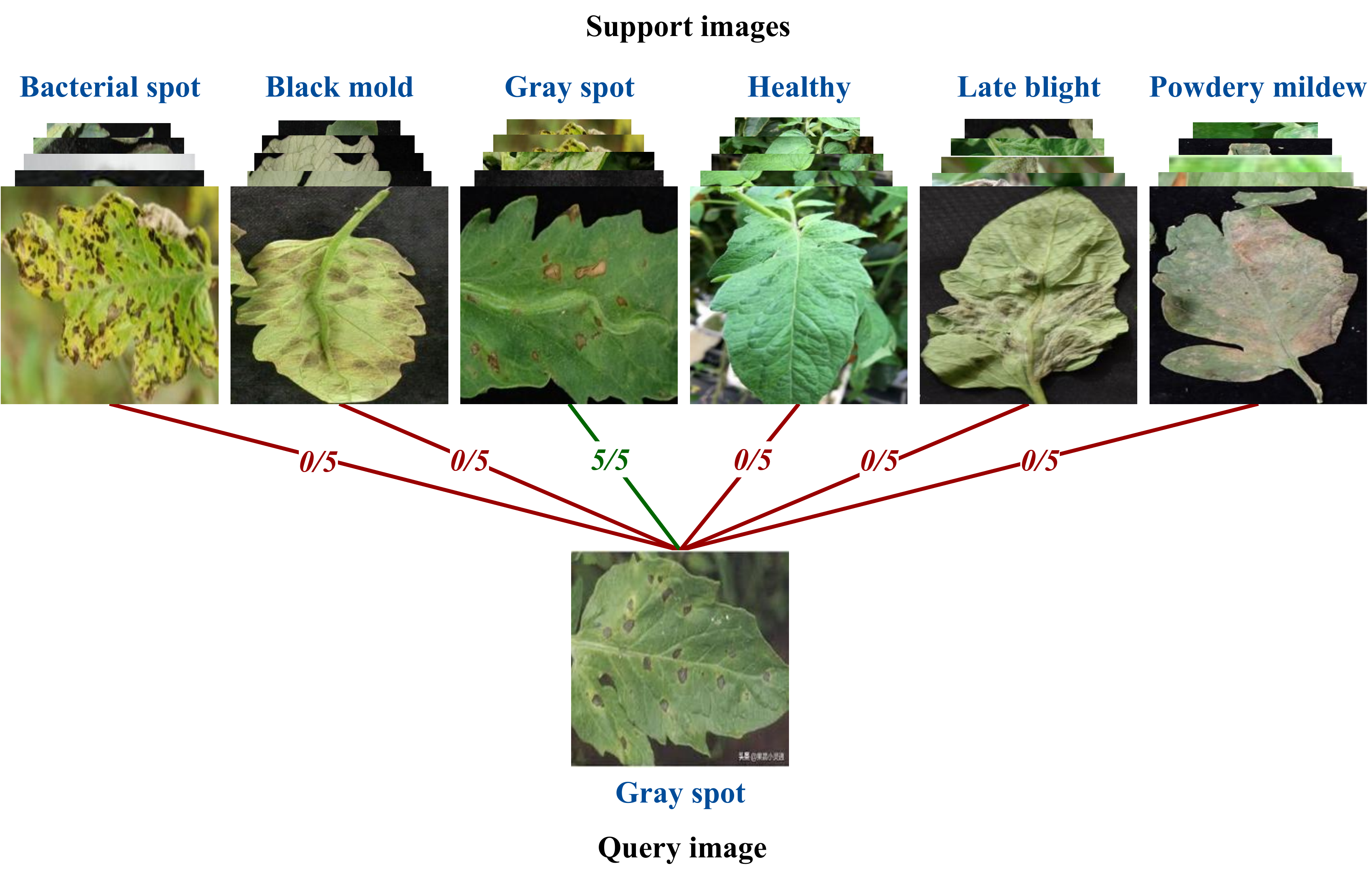}
	\caption{Querying a test sample from Gray spot disease class of the Taiwan dataset. The }
	\label{fig:querying}
\end{figure}

The testing framework illustrated in Figure \ref{fig:testing_framework} is exploited to obtain the testing accuracy of the proposed tomato disease recognition approach. Figure \ref{fig:querying} provides an example test scenario where a Gray spot test image is queried against the support images. The query image obtained smaller distances with all five Gray spot support images compared to the query images of other classes. Hence, using the majority voting scheme, the query image is correctly classified as the Gray spot diseased leaf on this occasion.

\begin{figure}
	\centering
	\includegraphics[width=\linewidth,scale=1.5]{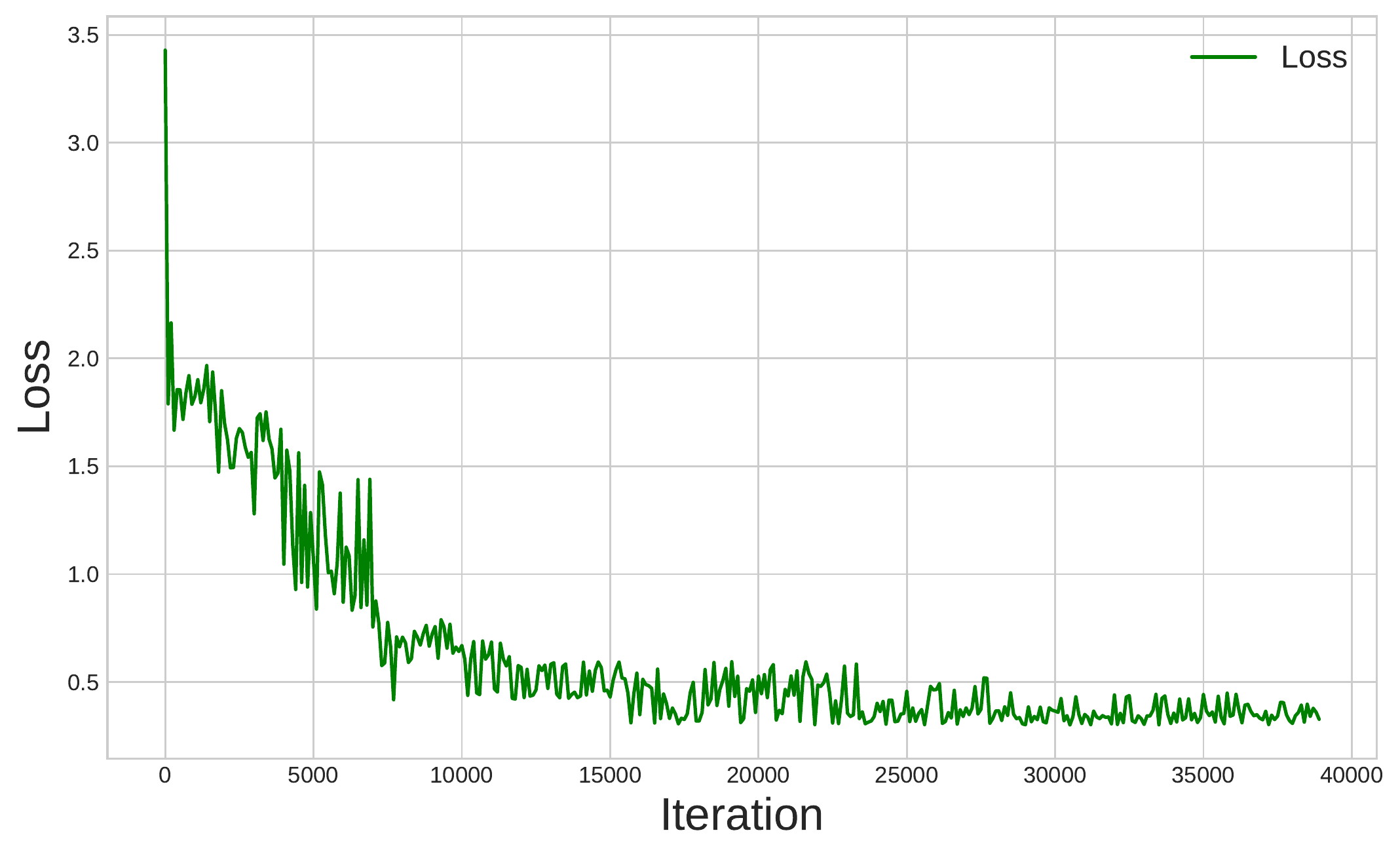}
	\caption{The loss observed for the model trained on the whole PlantVillage tomato leaf dataset.}
	\label{fig:loss}
\end{figure}

Figure \ref{fig:loss} presents the loss observed for the model trained with 100\% data of the PlantVillage tomato leaf dataset. The loss curve demonstrates that the model learned smoothly with given parameters. Table \ref{tab:inputsequence} summarizes the overall accuracy obtained for the proposed framework on the PlantVillage and both variations of the Taiwan tomato leaf disease datasets. The results are presented for the models trained with 100\%, 75\% and 50\% train data. The proposed framework trained with 100\% train set obtained the accuracy of 95.48\%, 96.14\% and 96.97\% on the original Taiwan dataset, Taiwan dataset with augmentation and the PlantVillage dataset, respectively. In all three cases, an increase in the accuracy is noted for the Taiwan dataset with augmentation than the original Taiwan dataset, which confirms that the augmentation enhances the performance of the proposed framework. Further, the overall accuracy is consistent even after reducing the training data to 75\% and 50\%, with a maximum 4\% accuracy drop recorded for the Taiwan dataset. For the PlantVillage tomato leaf dataset, only a 1.5\% drop is witnessed when reducing the training set by 50\%, which is convincing that the proposed framework can show good performance with small data.

\begin{table}
	\caption{Average accuracy achieved on the benchmark tomato leaf disease datasets. The results are separately recorded for the models trained with 50\%, 75\% and 100\% of the train set.}
	\begin{center}
		\begin{threeparttable}
			\begin{tabular}{|l|l|c|c|}
				\hline
				\multirow{2}{*}{\textbf{Databases\Tstrut}} & \multicolumn{3}{c|}{\textbf{Training Data \Tstrut}} \\
				\cline{2-4}
				&\textbf{100\%}	&\textbf{75\%}	&\textbf{50\%\Tstrut} \\
				\hline
				\textbf{Taiwan \Tstrut}
				&.9548&.9262&.9103\\
				
				\textbf{Taiwan /w augmentation \Tstrut}				
				&.9614&.9434&.9279\\
				
				\textbf{PlantVillage\Tstrut}	
				&.9697&.9609&.9546\\
				
				\hline
			\end{tabular}
			
			\begin{tablenotes}
				\small
				\item The accuracy presented for both variations of the Taiwan dataset is based on the dedicated test data provided by the authors.
			\end{tablenotes}
			
		\end{threeparttable}
	\end{center}
	
	\label{tab:inputsequence}
\end{table}

\begin{figure}
	\centering
	\includegraphics[width=\linewidth,scale=1.5]{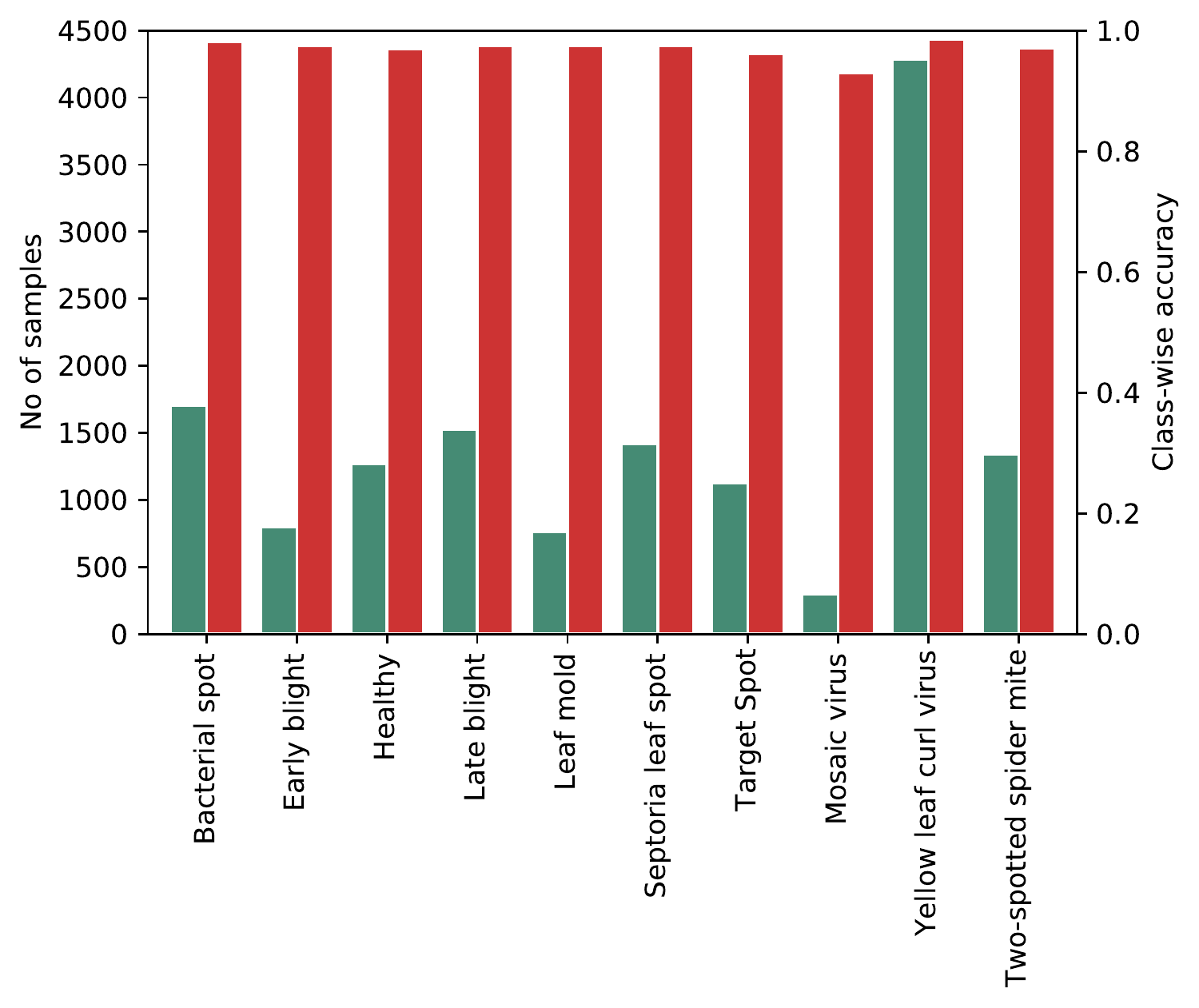}
	\caption{Performance evaluation with imbalanced data: per class samples and class-wise accuracies for the PlantVillage dataset are compared.}
	\label{fig:imbalanceddataresults}
\end{figure}

Figure \ref{fig:imbalanceddataresults} shows the number of samples and accuracies obtained for each class in the PlantVillage dataset. The results are recorded for the model trained on 100\% data. In each pair of bars, the left and right bars indicate the number of samples and the class-wise accuracies, respectively. As can be seen, the tomato subset of the PlantVillage dataset is highly imbalanced, where the Mosaic virus disease is the minority class with $299$ samples and Yellow leaf curl virus is the majority class with $4286$ samples (i.e., greater than $14\times$ of the minority class). However, the proposed framework obtained higher class-wise accuracies for minority classes. This demonstrates that the proposed framework can consistently perform well on imbalanced data.

\subsection{Comparison}
Two types of comparisons are performed to show the superiority of the proposed framework. In the first case, the backbone network of the proposed framework is replaced by other state-of-the-art networks. The rest of the procedures are kept identical, including the training and testing schemes. In the second case, some of the recently proposed tomato disease recognition methods are compared. In this case, the same training and testing environments are maintained for a fair evaluation.

Table \ref{tab:comparison_deepnetworks} provides the comparison results obtained for the proposed lightweight network and eleven state-of-the-art networks as the backbone of the proposed framework. In this comparison, both the heavy networks and lightweight networks are included. In order to perform a fair evaluation, all the backbones are trained from scratch. The proposed framework with the novel lightweight deep network achieved the best accuracy on all tomato datasets with 100\%, 75\% and 50\% data. In particular, the proposed framework with a novel lightweight deep network showed only a slight depreciation in the accuracy when the data was reduced to 75\% and 50\%. This is a promising feature as the performance of most of the existing state-of-the-art networks drops dramatically while reducing the training data. Notably, another lightweight architecture MobileNetv2, also consistently performed across all datasets with the second-best accuracy. However, its accuracy is significantly lower than the proposed framework integrated with the novel lightweight network.

\begin{table*}
	\caption{Comparison of the average accuracies obtained on the PlantVillage, Taiwan and Taiwan with augmentation tomato leaf disease datasets. The trainable parameters for each single stream network are also compared.}
	\begin{center}
		\begin{threeparttable}
			\begin{tabular}{|l|c|c|c|c|c|c|c|c|c|c|}
				\hline
				\multirow{2}{*}{\textbf{Deep Networks\Tstrut}} &
				\multirow{2}{*}{\textbf{Params\Tstrut}} & \multicolumn{3}{c|}{\textbf{PlantVillage \Tstrut}} &
				\multicolumn{3}{c|}{\textbf{Taiwan \Tstrut}} & \multicolumn{3}{c|}{\textbf{Taiwan /w augmentation \Tstrut}} \\
				\cline{3-11}
				&
				&\textbf{100\%}	&\textbf{75\%}	&\textbf{50\%\Tstrut} &\textbf{100\%}	&\textbf{75\%}	&\textbf{50\%\Tstrut}
				&\textbf{100\%}	&\textbf{75\%}	&\textbf{50\%\Tstrut}\\
				\hline
				\textbf{AlexNet \Tstrut}
				&61.1008 
				&.7128&.6812&.6137
				&.6896&.6039&.5527
				&.7064&.6788&.6438\\
				
				\textbf{VGG19 \Tstrut}
				&143.6672 
				&.7378&.7031&.6374
				&.6931&.6438&.6078
				&.7217&.7033&.6612\\
				
				\textbf{ResNet152 \Tstrut}
				&60.1928
				&.8316&.7812&.7079
				&.6018&.5178&.4538
				&.7423&.7164&.6891\\
				
				\textbf{DenseNet201 \Tstrut}
				&20.0139
				&.8134&.7521&.7215
				&.6135&.5730&.5137
				&.7239&.6433&.6138\\
				
				\textbf{Inceptionv3 \Tstrut}
				&27.1613	
				&.8436&.8127&.7513
				&.6847&.6037&.5678
				&.7869&.7519&.7238\\
				
				\textbf{SqeezeNet1\_1 \Tstrut}
				&1.2355
				&.7825&.7436&.7212
				&.6622&.6231&.5972
				&.7634&.7439&.7094\\
				
				\textbf{ShuffleNetv2 \Tstrut}
				&7.3940	
				&.7563&.7103&.6479
				&.7231&.6012&.5678
				&.7013&.6833&.6671\\
				
				\textbf{MobileNetv2 \Tstrut}
				&3.5049	
				&.9103&.8563&.8117
				&.8236&.7863&.7631
				&.8437&.8032&.7769\\
				
				\hline
				
				\textbf{Ours \Tstrut}
				&2.9629	
				&\textbf{.9697}&\textbf{.9609}&\textbf{.9546}
				&\textbf{.9548}&\textbf{.9262}&\textbf{.9103}
				&\textbf{.9614}&\textbf{.9434}&\textbf{.9279}\\
				
				\hline
			\end{tabular}
			
			\begin{tablenotes}
				\small
				\item The accuracy presented for both variations of the Taiwan dataset is based on the dedicated test data provided by the authors.
			\end{tablenotes}
			
		\end{threeparttable}
	\end{center}
	
	\label{tab:comparison_deepnetworks}
\end{table*}

The trainable parameters of each backbone network are compared in Table \ref{tab:comparison_deepnetworks}. The proposed lightweight network has nearly 2.9629 million trainable parameters, which is less than many state-of-the-art deep networks. SqueezeNet has the lowest trainable parameters among the competitors, with approximately 1.2355 million. Another lightweight network, namely ShuffleNet, has around 7.3940 million parameters. Except for these three lightweight networks, all other networks are very deep networks with larger trainable parameters. For instance, the VGG19 has nearly 143.6672 parameters to tune during the training, which is time-consuming and requires large training data. The proposed framework is faster and shows high performance with small data because the novel lightweight network integrated into the proposed framework has few parameters.

A comparison between the proposed framework and other existing tomato leaf disease recognition methods is given in Table \ref{tab:comparison_existingmethods}. The results presented for the compared methods are not directly taken from the respective papers as the experimental environment is entirely different. Hence, these methods are rerun on the tomato datasets under the same environment with 100\%, 75\% and 50\% data for comparison. As can be seen from the comparison, the proposed framework comprehensively outperformed existing tomato leaf disease recognition methods. The proposed framework showed a 13.16\%, 25.33\% and 14.05\% accuracy improvement over its competitor, \cite{astani2022diverse}, with 50\% data on the PlantVillage, Taiwan and Taiwan with augmentation tomato datasets, respectively. The proposed framework showed a higher accuracy improvement (25.33\%) with the Taiwan dataset, which demonstrates that the proposed framework can better recognize tomato leaf disease recognition with small data. 

\begin{table*}
	\caption{Comparison of existing tomato leaf disease recognition methods.}
	\begin{center}
		\begin{threeparttable}
			\begin{tabular}{|l|c|c|c|c|c|c|c|c|c|}
				\hline
				\multirow{2}{*}{\textbf{Methods\Tstrut}} &
				\multicolumn{3}{c|}{\textbf{PlantVillage \Tstrut}} &
				\multicolumn{3}{c|}{\textbf{Taiwan \Tstrut}} & \multicolumn{3}{c|}{\textbf{Taiwan /w augmentation \Tstrut}} \\
				\cline{2-10}
				&\textbf{100\%}	&\textbf{75\%}	&\textbf{50\%\Tstrut} &\textbf{100\%}	&\textbf{75\%}	&\textbf{50\%\Tstrut}
				&\textbf{100\%}	&\textbf{75\%}	&\textbf{50\%\Tstrut}\\
				
				\hline
				\textbf{Agarwal et al. \cite{agarwal2020toled} \Tstrut}
				&.8863&.8136&.7125
				&.6821&.6338&.5822
				&.8368&.7791&.7137\\
				
				\textbf{Chen et al. \cite{chen2020identification} \Tstrut}
				&.9012&.8633&.7473
				&.8036&.7191&.6361
				&.8672&.7736&.7236\\
				
				\textbf{Trivedi et al. \cite{trivedi2021early} \Tstrut}
				&.8463&.7966&.7138
				&.7120&.6731&.6133
				&.7463&.7038&.6477\\
				
				\textbf{Bhujel et al. \cite{bhujel2022lightweight} \Tstrut}
				&.9136&.8817&.7968
				&.7934&.7367&.6982
				&.8892&.8367&.7962\\
				
				\textbf{Astani et al. \cite{astani2022diverse} \Tstrut}
				&.9321&.9068&.8436
				&.8463&.7792&.7263
				&.9132&.8766&.8136\\
				
				\hline
				
				\textbf{Ours \Tstrut}
				&\textbf{.9697}&\textbf{.9609}&\textbf{.9546}
				&\textbf{.9548}&\textbf{.9262}&\textbf{.9103}
				&\textbf{.9614}&\textbf{.9434}&\textbf{.9279}\\
				
				\hline
			\end{tabular}
			
			\begin{tablenotes}
				\small
				\item The accuracy presented for both variations of the Taiwan dataset is based on the dedicated test data provided by the authors.
			\end{tablenotes}
			
		\end{threeparttable}
	\end{center}
	
	\label{tab:comparison_existingmethods}
\end{table*}


\section{Conclusion}
\label{sec:conclusion}
Timely recognition of tomato leaf diseases with small and imbalanced data is essential to overcome practical challenges related to tomato crop cultivation, such as unprecedented losses. So far, only a few existing methods have studied tomato leaf disease recognition. However, none of them has explored the small and imbalanced data problems in this domain. It is also important to develop a lightweight deep learning approach that can be deployed directly on resource constraint devices. Hence, in this work, a Siamese network-based lightweight framework is proposed to perform effective tomato leaf disease recognition. The proposed framework is integrated with a novel lightweight deep network as the backbone. The proposed framework, along with the novel testing procedure, is tested on three benchmark tomato leaf datasets, namely the PlantVillage, Taiwan and Taiwan with augmentation. The results demonstrated that the proposed framework effectively recognises tomato diseases with small and imbalanced training data. Further, the proposed framework's single stream network consists of 2.9629 trainable parameters, which is way fewer than the existing state-of-the-art deep networks. Comparative analysis confirms that the proposed tomato leaf disease recognition framework is superior to recently proposed methods. In the future, the proposed framework will be evaluated with the data collected in-the-wild as it is significant to test the in-field applicability.

\printcredits

\balance
\bibliographystyle{model1-num-names}

\bibliography{cas-refs}


\bio{figs/thuseethan}
\textbf{Selvarajah Thuseethan} received the B.Sc. degree from University of Jaffna, Sri Lanka, and the Ph.D. degree from Deakin University, Australia. He is currently a Probationary Lecturer with the Department of Computing and Information Systems, Sabaragamuwa University of Sri Lanka. He has authored over 25 refereed journals, book chapters, and conference articles. His research interest includes deep learning, emotion recognition, computer vision, and pattern recognition.
\endbio

\bio{figs/vicky}
\textbf{Palanisamy Vigneshwaran} received the B.Sc. degree from Vavuniya Campus of the University of Jaffna, Sri Lanka. He is currently a Probationary Lecturer with the Department of Computing and Information Systems, Sabaragamuwa University of Sri Lanka. He has authored over 10 refereed journals, book chapters, and conference articles. His research interest includes Deep learning, Software Process and Software Architecture.

\endbio

\bio{figs/charles}
\textbf{Joseph Charles} received the B.Sc. degree from Sabaragamuwa University of Sri Lanka, Sri Lanka. He is currently a Probationary Lecturer with the Department of Physical Sciences and Technology, Sabaragamuwa University of Sri Lanka. He has authored over 10 refereed journals, book chapters, and conference articles. His research interest includes deep learning, emotion recognition, audio data mining, and pattern recognition.
\endbio

\bio{figs/chathrie}
\textbf{Chathrie Wimalasooriya} is currently pursuing the Ph.D. degree at University of Otago, New Zealand. She received the ME degree from University of Canterbury, New Zealand and B.Sc. degree from Sabaragamuwa University, Sri Lanka. She has been a lecturer in the Department of Computing and Information Systems and Sabaragamuwa University since 2015. Her research interests are empirical software engineering, software maintenance and machine learning applications.
\endbio
\end{document}